\documentclass{bmvc2k}
\usepackage{kotex}
\usepackage{multirow, tabularx}
\usepackage{booktabs}
\usepackage{adjustbox}
\usepackage{graphicx}
\usepackage{array}
\usepackage{dsfont}
\usepackage{bm}
\usepackage{amsmath}
\usepackage{xcolor}

\title{MC-GAN: Multi-conditional Generative Adversarial Network for Image Synthesis
}

\runninghead{Student, Prof, Collaborator}{BMVC Author Guidelines}

\addauthor{Hyojin Park}{wolfrun@snu.ac.kr}{1}
\addauthor{YoungJoon Yoo}{youngjoon.yoo@navercorp.com}{2}
\addauthor{Nojun Kwak}{nojunk@snu.ac.kr}{1}

\addinstitution{
 Department of Transdisciplinary Studies, \\
 Seoul National University, \\
 Republic of Korea.
}
\addinstitution{
 CLOVA AI Research, Naver Corp., \\
 Republic of Korea.
}

\runninghead{H. Park, Y. Yoo and N. Kwak}{MC-GAN}

\def\ie{\emph{i.e}\bmvaOneDot}
\def\eg{\emph{e.g}\bmvaOneDot}

\newif\ifdraft\draftfalse
\newif\ifdraft\drafttrue
\ifdraft

\else

}
\fi

\begin{document}
\maketitle

\begin{abstract}
In this paper, we introduce a new method for generating an object image from text attributes on a desired location, when the base image is given. 
One step further to the existing studies on text-to-image generation mainly focusing on the object's appearance, the proposed method aims to generate an object image preserving the given background information, which is the first attempt in this field.
To tackle the problem, we propose a multi-conditional GAN (MC-GAN) which controls both the object and background information jointly. 
As a core component of MC-GAN, we propose a synthesis block 
which disentangles the object and background information in the training stage.
This block enables MC-GAN to generate a realistic object image with the desired background by controlling the amount of the background information from the given base image using the foreground information from the text attributes. 
%
From the experiments with Caltech-200 bird and Oxford-102 flower datasets, we show that our model is able to generate photo-realistic images with a 
resolution of $128 \times 128$. The source code of MC-GAN is released.\footnote{\url{https://github.com/HYOJINPARK/MC_GAN}}

\end{abstract}
\section{Introduction}
\label{sec:intro}

Recent studies on generative adversarial networks (GAN)~\cite{goodfellow2014generative,mao2017least,chen2016infogan,radford2016unsupervised,gregor2015draw} have achieved impressive success in generating images.
However, since most of the generated images through GAN do not exactly satisfy the users' expectation, people use auxiliary information in various forms 
such as base images and texts 
as main cues for controlling the generated images.
In this line of research, it is actively studied to create images under specific conditions such as transferring the style of an image  \cite{choi2017stargan,kim2017disco,CycleGAN2017,DBLP:conf/nips/MaJSSTG17,chen2017photographic} or generating an image 
based on text description \cite{Han17stackgan2,zhang2017stackgan,reed2016generative,mansimov16text2image,reed2016pixelcnn}.





Among these, text-to-image generation is meaningful in that it can fine-tune the generated image through the guide of text description. Although GAN can be applied to diverse text to image generation works, most of the applications are focused on controlling the shape and texture of the foreground and relatively less attention has been paid to the background.
For example, \citet{reed2016learning} created images from the text containing information on the appearance of the object to generate. The method can generate a target image in a given location or with a specified pose. However, they also cannot control the background.
There are some works \cite{dong2017semantic,DBLP:conf/nips/MaJSSTG17} that have considered the background. 
\citet{dong2017semantic} considered multi-modal conditions based both on image and text and can change the base image according to the text description. However, the method has a restriction that a similar object to the generated one should be in the base image. Thus, it can be considered as a style transfer problem.
%
 \citet{DBLP:conf/nips/MaJSSTG17} also solved the multi-modal style transfer problem from a reference person image to a target pose. They kept the background and changed the reference person's pose to a target pose. 


\begin{figure}
  \centering
    \begin{tabular}{m{0.2\textwidth} m{0.8\textwidth}}
    \centering\citet{reed2016learning}
    & \bmvaHangBox{\includegraphics[width=0.7\textwidth]{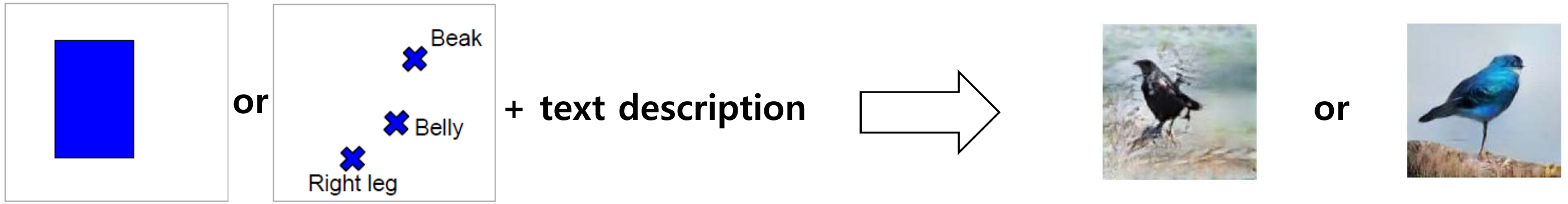}} \\
    \centering\citet{dong2017semantic} 
    & \bmvaHangBox{\includegraphics[width=0.7\textwidth]{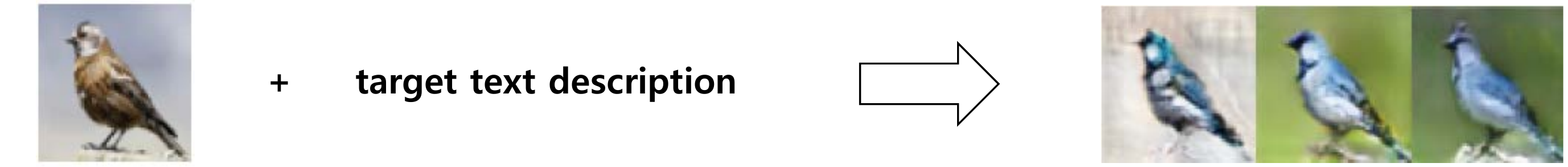}} \\
    \centering Ours  
    & \bmvaHangBox{\includegraphics[width=0.7\textwidth]{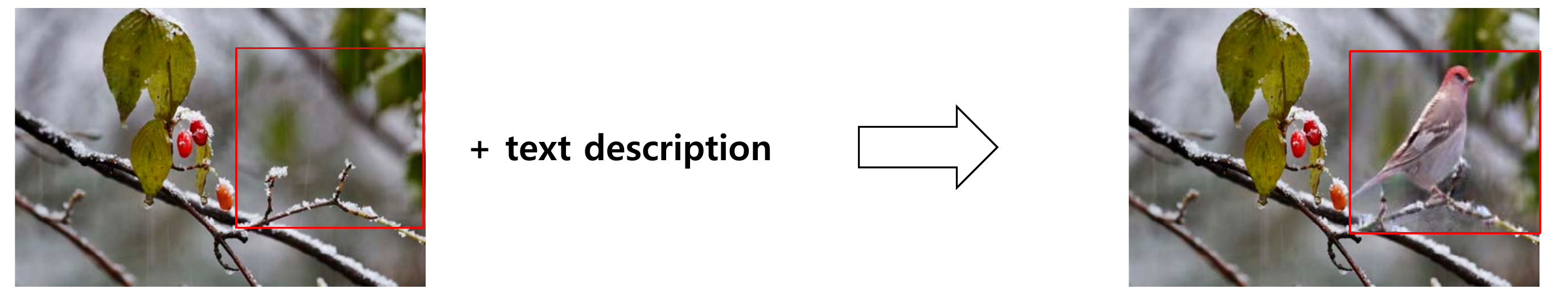}} \\
    \end{tabular}%
        
   \caption{Comparison of different multi-modal conditional GAN problems. We try to synthesize an image based on a base background image with a specified location for an object and a text description on the object. The images on the right are from the respective algorithms.}
  \label{fig:base}%
  
  \vspace{-3mm}
 
\end{figure}%

In this paper, we define a novel problem of conditional GAN which generates a new image by synthesizing the background of an original base image and a new object described by the text description in a specific location.
Different from the existing works \cite{reed2016learning,dong2017semantic}, we aim to draw a target object on a base image that does not contain similar objects.
To the best of our knowledge, our research is the first attempt to synthesize a target image by combining the background of an original image and a text-described foreground object.
As shown in Fig. \ref{fig:base}, our approach is different from other studies that try to create a random image at a desired location  \cite{reed2016learning} or to change the foreground style \cite{dong2017semantic} in that we want to independently apply separate foreground and background conditions for image synthesis.
This problem is not trivial because the generated foreground object and the background from the base image should be smoothly mixed with a plausible pose and a layout. 

To tackle this problem of image synthesis, we introduce a new architecture of multi-conditional GAN (MC-GAN) using a synthesis block which acts like a pixel-wise gating function controlling
the amount of information from the base background image by the help of the the text description for a foreground object.
%
With the help of this synthesis block, MC-GAN can generate a natural image of an object described by the text in a specified location with the desired background.
To show the effectiveness of our method, we trained MC-GAN using the Caltech-200 bird dataset \cite{WahCUB2002011} and the Oxford-102 flower dataset \cite{Nilsback08,Nilsback07Seg} and compared the performances with those of a baseline model \cite{dong2017semantic}. 


\begin{figure}
  \centering  
    {\bmvaHangBox{\includegraphics[width=0.95\textwidth]{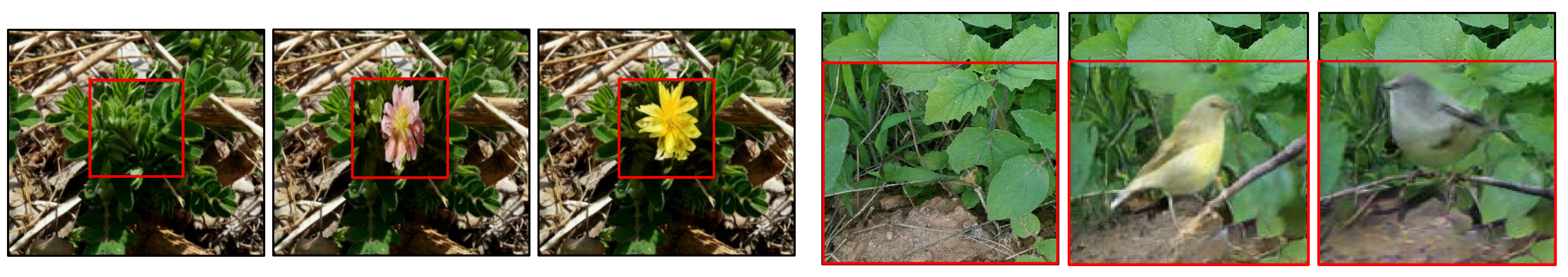}}}   
     \caption{The examples of synthesized images using bird and flower datasets}
  \label{fig:inference}%
  \vspace{-3mm}
\end{figure}%

Our main contributions can be summarized as follows: (1) We define a novel multi-modal conditional synthesis problem using base image, text and location. (2) To handle complex multi-modal conditions in GAN, we suggest a new architecture of MC-GAN using synthesis blocks. (3) 
The proposed architecture was shown to generate a plausible natural scene as shown in Fig. \ref{fig:base} and \ref{fig:inference}
by training publicly available data regardless of whether the base image contains a similar object to create or not.

\section{Related Work}
\label{related_work}

Among the diverse variants of GAN~\cite{CycleGAN2017,choi2017stargan,choi2017stargan,reed2016generative,wang2018pix2pixHD}, we can mainly categorize the studies into three large groups: 1) the style transfer problem 2) the text-to-image problem 3) the multi-modal conditional problem.

\noindent
\textbf{Style Transfer: }
The style transfer problem uses an image as input and converts the foreground to a different style.
\citet{CycleGAN2017, choi2017stargan} and \citet{kim2017disco} transfered images to a different domain style, such as a smile face to an angry face, or a handbag to shoes.
In addition to these applications, some works created a real image from a map of segmentation label \cite{wang2018pix2pixHD, chen2017photographic}, or used a map of part location in combination with an original image of a person to generate images of a person with different poses \cite{DBLP:conf/nips/MaJSSTG17}. 

\noindent
\textbf{Text-to-Image: }
The text-to-image problem uses text description as an input to generate an image. 
It has great advantages over other methods
in that it can easily generate an image with the attributes that a user really wants, because text can express detailed high-level information 
on the appearance of an object with detailed attributions. The raw text is usually 
embedded according to the method in \cite{reed2016txtemb} which uses a hybrid of CNN (convolutional neural network) and RNN (recurrent neural network) structure. 
\citet{reed2016generative} proposed a novel text-to-image generation model, and \citet{zhang2017stackgan,Han17stackgan2} improved the image quality later by stacking multiple GANs.

\noindent
\textbf{Multi-modal Conditional Image Generation: }
A multi-modal conditional problem is to create images satisfying multiple input conditions in different modalities 
such as a pair of (image, location) or (image, text). 
\citet{reed2016learning} provided a desired object position by a bounding box or a set of object part locations by points in an empty image in combination with the text description to generate an object image (see Fig. \ref{fig:base}). 
\citet{dong2017semantic} used both an image and a text as input to GAN for image generation (see also Fig. \ref{fig:base}). 
They intended to keep the image part irrelevant to the text and to change the style of the object contained in the base image based on the text description.
Although \cite{reed2016learning} is similar to our work in terms of using the location information, our method generates an object with an appropriate pose by understanding the semantic information of the background image automatically. 
Compared to our method, the method of using parts' locations in \cite{reed2016learning}, which requires a user to select parts' locations, is somewhat unnatural and time-consuming. 
%
However, bounding box condition in \cite{reed2016learning} is similar to our problem, thus it can be said that our study partially includes the problem defined in \cite{reed2016learning}. 
The method in \cite{dong2017semantic} also uses image and text conditions together. 
However, our study does not have a restriction that the same kind of object to generate must be in the base image. 
In other words, ours does not change the style of an already existing object but synthesize a new object with a slightly but properly changed background. (See the last row of Fig. \ref{fig:base} and  \ref{fig:inference}.)


\section{Methodology}
\label{method}
\begin{figure}[t]
\centering
\begin{tabular}{c}
\bmvaHangBox{
\includegraphics[width=0.9\linewidth]{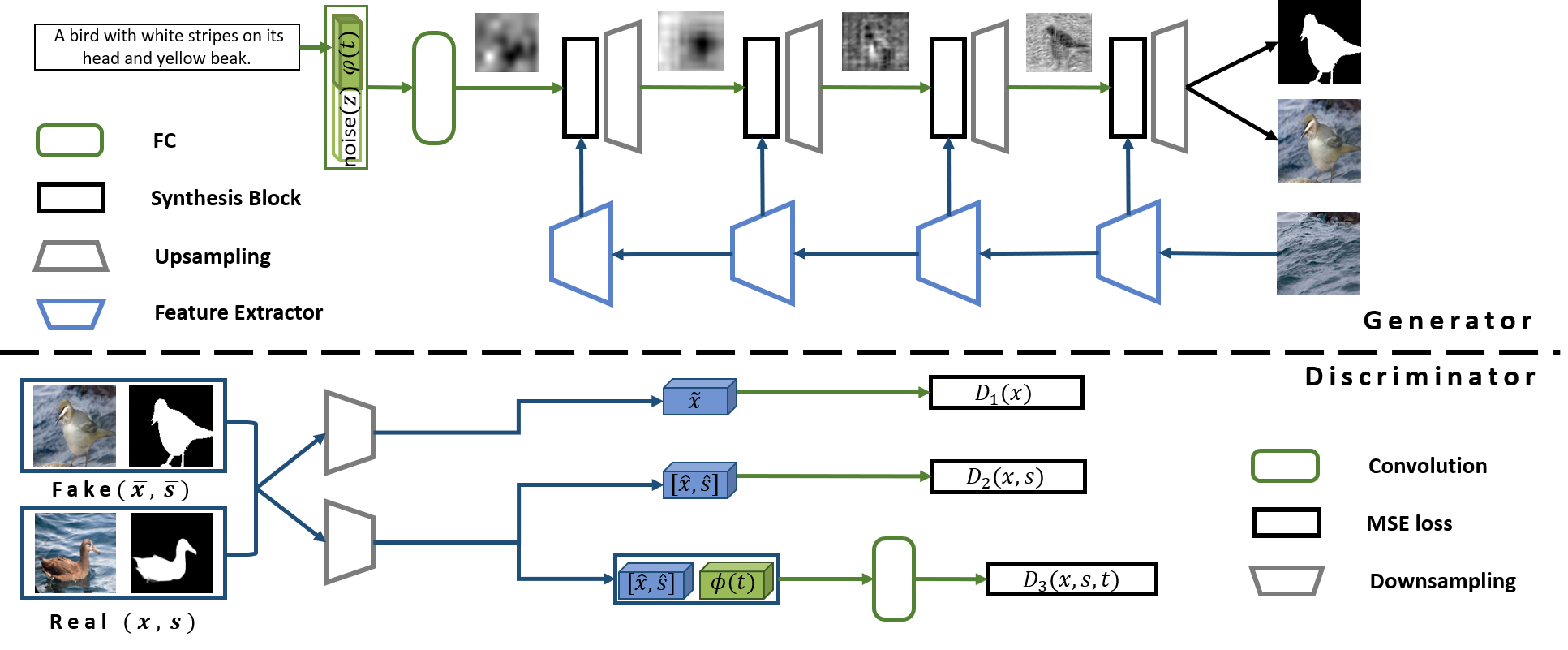}
}
\end{tabular}
\caption{Overall structure of MC-GAN. The generator uses text description as well as the base image to create an image of the object as well as a segmentation mask. Using the image and the segmentation mask, the discriminator distinguishes whether the input is real or not.
}
\vspace{-3mm}
\label{fig:arch}
\end{figure}

Fig. \ref{fig:arch} represents the overall structure of the proposed MC-GAN.
The generator of MC-GAN firstly encodes the input text sentence $t$ into the text embedding $\varphi(t)$ using the method in \cite{reed2016txt_emb}.
As in \cite{zhang2017stackgan} and \cite{Han17stackgan2}, $\varphi(t)$ is concatenated with a noise vector $z$ to which fully connected (FC) layers are applied to constitute a seed feature map.
After then, we use a series of synthesis blocks whose inputs are the seed feature map, which in combination with the image features from the background image generates an output image and a segmentation mask. 
The synthesis blocks are used
to prevent overlapping between the generated object and the background. 
In Section \ref{sec:synthesis-block}, we describe the characteristics of the proposed synthesis block in more detail and
we introduce the detailed explanation of the model structure and the training strategy in 
Section \ref{sec:arch},

\subsection{Synthesis Block}
\label{sec:synthesis-block}

\begin{figure}[t]
  \centering
  \begin{adjustbox}{width=\textwidth}
    \begin{tabular}{c m{0.4\textwidth}}
     \multicolumn{1}{c}{\multirow{5}[0]{*}{ 
    \bmvaHangBox{\includegraphics[width=0.6\textwidth]{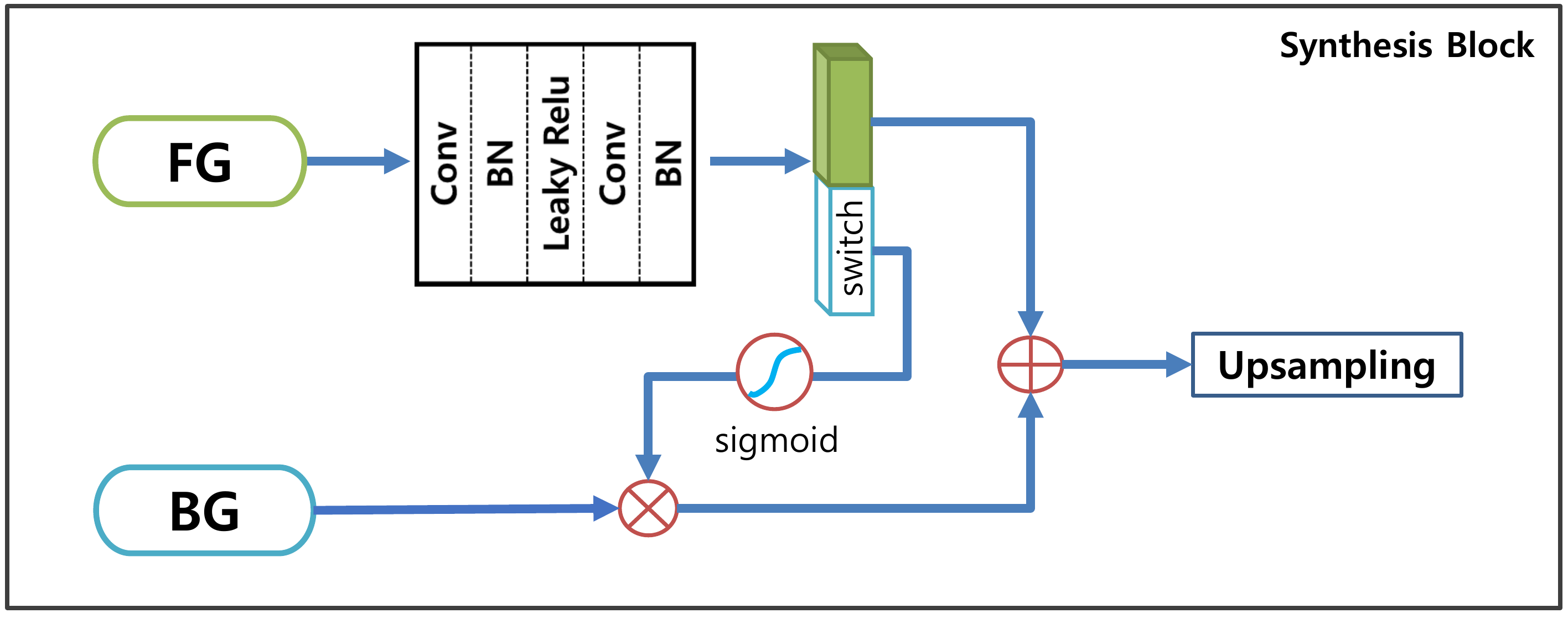}}}}  
    &\bmvaHangBox{\includegraphics[width=0.4\textwidth]
{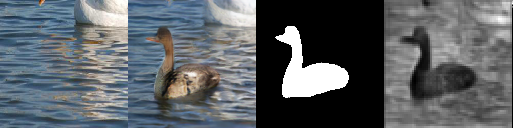}}  \\  

          & \tiny{A larger sized bird, solid white in color, with subtle speckles of brown, that has a long reddish beak with a bump.}  \\      
 &\multicolumn{1}{c}{(b)} \\
    &  \bmvaHangBox{\includegraphics[width=0.4\textwidth]
{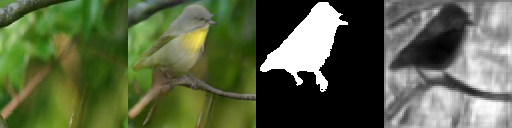}} \\
          & \tiny{A small mostly grey bird with a yellow crown, black eyerings, white cheeks, dark grey breast and a bright yellow wingbar on grey wings.}\\
           (a)   & \multicolumn{1}{c}{(c)} \\
    \end{tabular}%
  \end{adjustbox}  
  \caption{The structure of synthesis block (a) and the examples of generated image (b) (c)}
  \vspace{-3mm}
  \label{fig:synthesis-block}  
\end{figure}%

Fig. \ref{fig:synthesis-block} (a) describes the framework of the proposed synthesis block. In the synthesis block, the background (BG) feature is extracted from the given image without non-linear function (\ie only using convolution and batch normalization (BN)) \cite{ioffe2015batch} and the foreground (FG) feature is the feature map from the previous layer.
As shown in the figure, the BG feature is controlled by multiplying it with an activated switch feature map, a fraction of the FG feature map.
The sizes of the BG and FG feature maps are the same, and the depth of FG feature map is doubled as it passes through the convolution layers. 
A half of the doubled feature map denoted as switch in the figure is used as an input to the sigmoid function, while the other half is forwarded to generate a larger FG feature map for the next synthesis block. 
The switch determines what amount of BG information should be retained for the next synthesis block. 
After switch feature map is activated by the sigmoid function, it is multiplied element-wise with the BG feature map for the suppression of background information where the object is to be generated. 
Finally, the spatial dimension of feature map is doubled by the upsampling layer after element-wise addition of the suppressed BG feature map with the FG feature map. 
Because the MC-GAN has $\ell_1$ loss comparing the background of the created image with that of the base image, the switch has an effect of suppressing the base image in the object area and mimicking the base image in the background. 
Therefore, a visualized switch map has the opposite concept to that of the segmentation mask.

Fig \ref{fig:synthesis-block} (b) and (c) show a couple of output examples. From left to right they are 1) a cropped background image from a specific location, 2) the generated image, 3) the generated mask and 4) the switch feature map from the final synthesis block. A close look at the switch feature map in Fig \ref{fig:synthesis-block} (b) shows that it does not change the original background much because the object naturally goes with the background.
Thus, the background region is highly activated in the switch while the object region is suppressed (see the last column).
On the other hand, in in Fig \ref{fig:synthesis-block} (c), because a picture of bird standing on the air is unnatural,
the generator adds a branch in the figure.
At this time, since the original background should not overlap with the newly generated branch, the background information of the new branch area deactivates the corresponding area of the switch map.

\subsection{Network Design and Loss Function}
\label{sec:arch}
 
MC-GAN encodes the input text sentence $t$ using the method in \cite{reed2016txt_emb}. However, the vector $\varphi(t)$ generated by this method lies on a high dimensional manifold, while the number of available data is relatively small. \citet{zhang2017stackgan,Han17stackgan2} pointed out this problem and proposed the \textit{conditioning augmentation} method and used fully connected layers to make an initial seed feature map from the text embedding $\varphi(t)$ and a noise vector $z$.
Here, we follow this method of creating the initial seed feature map as in \cite{zhang2017stackgan,Han17stackgan2}. 
A cropped region from the base image as well as the text sentence is inputted to MC-GAN. 
The spatial size of the input image is $ W \times H $ and the number of synthesis blocks is $ N $. The size of the seed feature map after fully connected layer is $1024 \times\frac{W}{2^N} \times \frac{H}{2^N}$ and the resolution is doubled with each pass through the upsampling layer. 
The upsampling layer uses the nearest neighborhood method to double the resolution and
a $3\times3$ convolution is applied with BN and ReLU activation to improve the quality of the image.
The number of channels is halved for each block of upstream.
Conversely, the method of creating an image feature (downstream) does not use any non-linear function, and each step consists of a $3\times3$ convolution layer with BN. 
At each step of downstream, the spatial resolution is halved using stride 2 and the number of channels is doubled. By combining the upstream and downstream, the final $ 64\times W \times H $ feature map is obtained which is converted into 4 channels (3 for RGB and 1 for segmentation masks).

The discriminator takes a tuple of image-mask-text ($x$-$s$-$t$) as an input. The convolution followed by BN and Leaky ReLU downsamples the image and the mask into an image code $\tilde{x}$ and an image-mask code $[\hat{x}, \hat{s}]$ separately both of which have a resolution of
$ \frac{W}{2^N}\times\frac{H}{2^N} $. 
%
%
The image-mask code is concatenated with the replicated text code $\phi(t)$, which is obtained by the \textit{conditioning augmentation technique} \cite{zhang2017stackgan,Han17stackgan2} using the text embedding $\varphi(t)$. 
We apply a convolution layer to the associated image-mask-text code, then perform BN and Leaky ReLU activation to reduce the dimension.
The image code $\tilde{x}$, image-mask code $[\hat{x},\hat{s}]$, and image-mask-text code $[\bar{x},\bar{s}, \bar{\phi}(t)]$ are trained by the method proposed in \citet{mao2017least}. In our case, the discriminator learns the following four types of input tuples. 

\vspace{2mm}
\resizebox{0.95\linewidth}{!}{
    \begin{tabular}{ll|l}
    1)    & image1$(x_1)$, mask1$(s_1)$, text1$(t_1)$ & real image with matching mask and text \\
    2)    & image1$(x_1)$, mask1$(s_1)$, text2$(t_2)$ & real image with matching mask but mismatching text \\
    3)    & image1$(x_1)$, mask2$(s_2)$, text1$(t_1)$ & real image with mismatching mask but matching text  \\
    4)    & fake image$({x}_g)$, fake mask$({s}_g)$, text$(t)$ & generated image and mask with input text \\
    \end{tabular}%
}
\vspace{2mm}
\\
\noindent 
Here, the subscript indicates whether the tuple matches or not. (\eg the tuple ($x_1, s_2, t_1$) means that the image matches with the text but the segmentation mask is mismatched.)
Using the four types of tuples, the discriminator loss function for the output $D_3$ becomes
\begin{equation}
\begin{split}
L_{D_3}  = &\mathds{E}_{x_1,s_1,t_1 \sim p_{d}} [ (D_3({x_1},{s_1},t_1)-1)^2 ] 
 + \mathds{E}_{x_1,s_1,t_2\sim p_{d}} [ {D_3({x_1},{s_1},t_2)}^2 ] \\
 &+ \mathds{E}_{{x_1},{s_2},t_1 \sim p_{d}} [ {D_3({x_1},{s_2}, t_1)}^2 ]
 + \mathds{E}_{x_g,s_g \sim p_{G(b,t,z)}, \ t\sim p_{d}} 
 [ D_3(x_g, s_g, t)^2 ].
\end{split}
\end{equation}
Here, $p_{d}$ and $p_G$ denote the distributions of real and generated data respectively.
$G(b,t,z)$ means the output of the generator using the base image $b$, text $t$ and noise $z$.
The first term enforces the discriminator to output 1 for the true input (type 1), the second term tries to distinguish mismatching texts (type 2), the third term for distinguishing false masks (type 3), and finally the last term is to distinguish the fake image and mask from the real one. 
Likewise, the loss functions for $D_1$ and $D_2$ become 
\begin{equation}
\begin{split}
L_{D_2}  &= \mathds{E}_{x_1,s_1 \sim p_{d}} [ (D_2({x_1},{s_1})-1)^2 ] 
 + \mathds{E}_{x_1,s_2\sim p_{d}} [ {D_2({x_1},{s_2})}^2 ]  + \mathds{E}_{ x_g,s_g \sim p_{G}} 
 [ D_2(x_g, s_g)^2 ] \\
L_{D_1}  &= \mathds{E}_{x_1 \sim p_{d}} [ (D_1(x_1)-1)^2 ] 
 + \mathds{E}_{ x_g \sim p_{G}} 
 [ D_1(x_g)^2 ] 
\end{split} 
\end{equation}


In the training of the generator, to the general loss term of GAN, regularization terms for the \textit{conditioning augmentation} and the background reconstruction are added as follows:
\begin{equation}
\begin{split}
L_G  =& \mathds{E}_{x,s \sim p_{G(b,t,z)}, \ t \sim p_d}
[ ({D_1(x)-1)})^2  + 
 ({D_2(x,s)-1)})^2  +
 ({D_3(x,s,t)-1)})^2  \\
 &+ \lambda_{1} D_{KL} ( \mathcal{N}(\mu(\phi(t)), \Sigma(\phi(t)) ) ||  \mathcal{N}(0, I))  +  
\lambda_2 ||(x \odot ( f \circ s)) - 
({b} \odot (f \circ s))||_1 ].
\label{eq:g_loss}
\end{split}
\end{equation}
Here, $\mu(\phi(t))$ and $ \Sigma(\phi(t))$ are the mean and the diagonal convariance matrix from the text embedding by \textit{conditioning augmentation} and the KL divergence loss term is used as in \cite{zhang2017stackgan,Han17stackgan2}.
The last term, the background reconstruction loss, affects the feature extraction of the base image and suggests the activation of the switch determining which part should be taken for synthesis. The operator $f \circ$ denotes the morphological erode operation to the mask $s$ for smoothing and $\odot$ is element-wise multiplication.
The areas in the fake image $x$ and the base image $b$ excluding the object part are taken and trained by the $\ell_1$ loss. 



\section{Experiment}
\label{Exp}

\subsection{Dataset and training details}
\label{data}
We validated the proposed algorithm using the publicly available Caltech-200 bird \cite{WahCUB2002011} and Oxford-102 flower \cite{Nilsback08} datasets.
For comparison, in addition to several ablation methods, we tested recent \citet{dong2017semantic}'s work which uses image and text based multi-modal conditions. 
\citet{reed2016learning} also used multi-modal conditions for generating images but they did not consider image condition and both \cite{dong2017semantic} and \cite{reed2016learning} are commonly based on the method in \citet{reed2016generative}. Thus, only the work of \citet{dong2017semantic} was compared.

Caltech-200 bird dataset consists of 200 categories of bird images (150 categories for train and 50 categories for test), and gives ground-truth segmentation mask maps for all the 11,788 bird images.
For the text attributes, we used the captions from \citet{reed2016txtemb} which contains $10$ captions for each image.  
The captions describe the attributes of a bird such as appearance and colors. 
For the background image, we cropped the image patches from the Caltech-200 bird dataset excluding birds by using the segmentation mask. Separate sets of background images were used for training and test.
 
Oxford-102 flower dataset includes $102$ categories of $8,189$ flower images.
The dataset is divide into a training set with $82$ categories and a test set with $20$ categories.
To achieve the ground truth segmentation mask, we used a segmentation method of \citet{Nilsback07Seg}.
For the background images, we crawled 1,352 images (1,217 for training and 135 for testing) from the web with keywords 'flower leaf' or 'flower foliage'.
The captions of the flower images from \cite{reed2016txtemb} were used, which describes the shape and colors of the flowers.

In the training, an initial learning rate of 0.0002 and Adam optimization \cite{kingma2015adam} with a momentum of $0.5$ were used.  
To generate $128\times128$ bird images, we used a batchsize of $32$ and trained the network for $1,200 \sim 1,500$ epochs.
For the flower dataset, we trained $900 \sim 1,200$ epochs. 
We set $\lambda_1 = 2$ for experiments on both datasets while $\lambda_2$ was set to 15 for the bird and 30 for the flower.  
We also used an image augmentation technique including flipping, zooming and cropping randomly. 
The size of the seed feature map was $8\times8$ and 4 synthesis blocks were used to generate an $128\times128$ image.

\subsection{Comparison with the Baseline Method}
\label{compare}
\begin{figure}
  \centering  
    {\bmvaHangBox{\includegraphics[width=\textwidth]{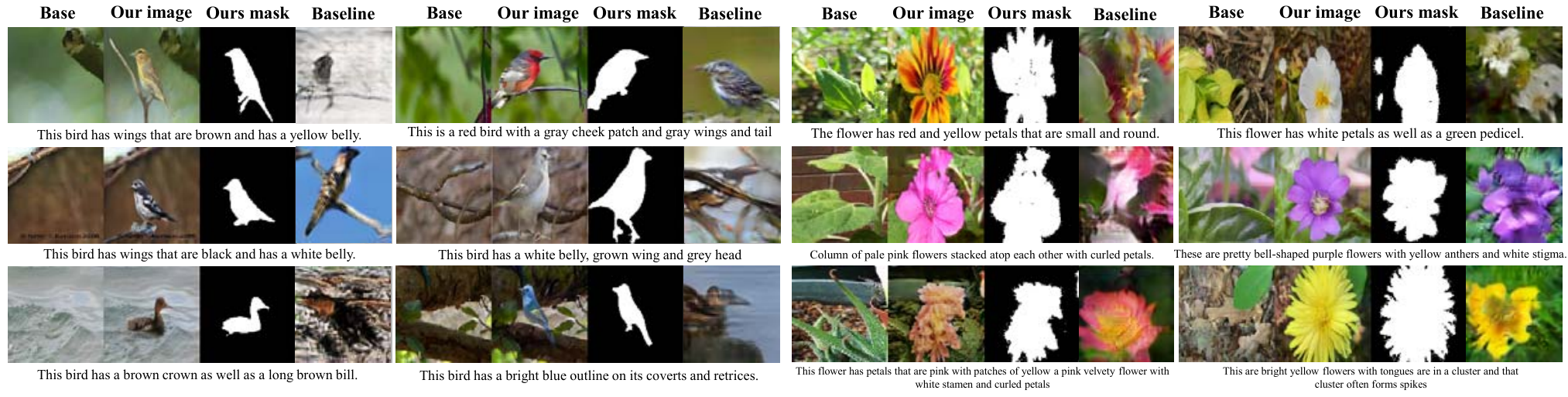}}}   
     \caption{Examples of multi-modal image synthesis. From left to right: Base image, the synthesized image and mask from MC-GAN ($128\times 128$) and baseline method \cite{dong2017semantic} ($64\times 64$)
     }   
  \label{fig:MC}%
  \vspace{-3mm}
\end{figure}%
We compared our method with the baseline \cite{dong2017semantic}, which is also an image-text multi-conditonal GAN, for our new synthesis problem. Originally, \citet{dong2017semantic} reduced the learning rate by 0.5 in every 100 epochs and trained until 600 epochs. 
In this experiment, we decreased the learning rate in every 200 epochs, and trained the network with the same number of epochs, for fair comparison.
Fig. \ref{fig:MC} shows some examples of generated images from the proposed MC-GAN and the baseline work of \cite{dong2017semantic}.
From the figures, we confirmed that the results from \cite{dong2017semantic} did not preserve the background information, or only generated background images without the target object.

\subsection{Comparison using the Synthesis Problem in \cite{dong2017semantic} }
\label{other-synthesis}
Here, we show that the synthesis block solves the multi-modal conditional problem more reliably than the baseline \cite{dong2017semantic}. 
The semantic synthesis problem of \citet{dong2017semantic} aims to keep the features of input images that are irrelevant to target text descriptions and to transfer the relevant part of the input image to the one that matches the target text description.  
We used the same text embedding method without segmentation mask, and we reduced the learning rate by 0.5 for every 100 epochs until 600 epochs under the same condition as \cite{dong2017semantic}. 
The proposed structure of MC-GAN is applied to the generator and  discriminator networks, but only the image-text pair loss is used for the discriminator as in \cite{dong2017semantic}. 

\begin{figure}
  \centering
    \bmvaHangBox{\includegraphics[width=\textwidth]{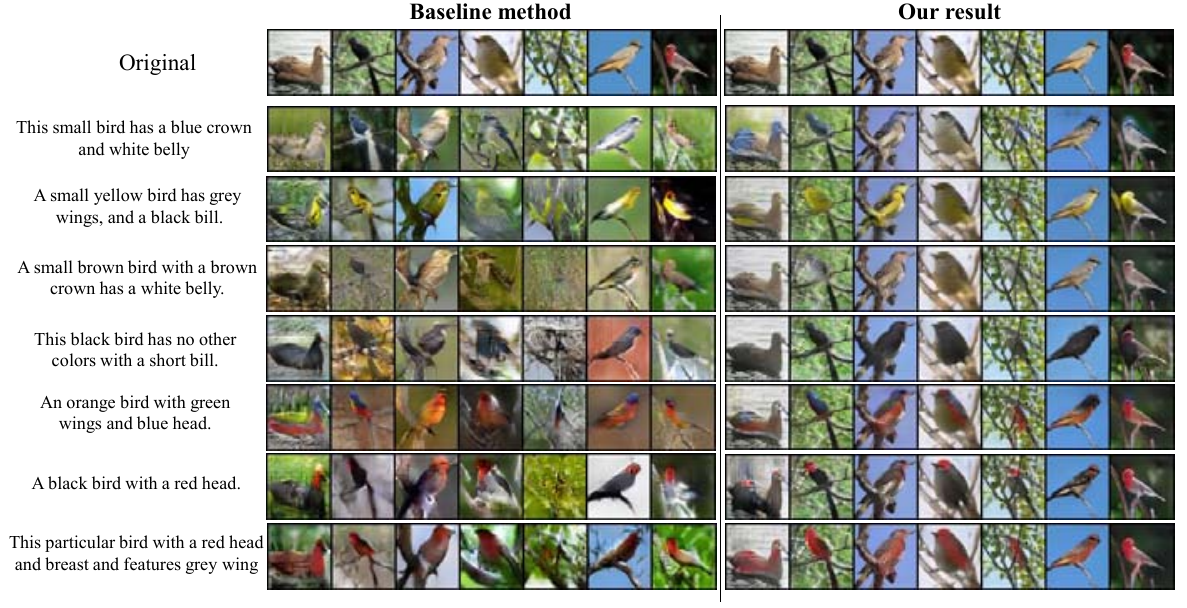}} 
    \caption{Example results for the image synthesis problem defined in \cite{dong2017semantic}. The baseline method \cite{dong2017semantic} and our method (MC-GAN without mask) on Caltech-200 bird test dataset}
  \label{fig:zeroshot}%
  \vspace{-3mm}
\end{figure}

Fig. \ref{fig:zeroshot} shows some examples of baseline method and ours. 
\citet{dong2017semantic} worked well if the background image is not complicated (column 3 and 6), but if the background is complex (column 5), it fails to generate a plausible object. Even though the object was generated well, the irrelevant part of image was also changed a lot. 
On the other hand, our generator using the proposed synthesis block stably maintained the shape of the object and the texture of the background even in a complex background, and the irrelevant background part to the text description rarely changes.
Although the segmentation mask is not used in the training, image features were provided to each synthesis block to keep the background and the shape of the object.
Therefore, the background part which was irrelevant to the target texts was maintained and at least the shape was not changed strangely even if the color of the object changed as the text description.

\begin{figure}
  \centering  
    {\bmvaHangBox{\includegraphics[width=0.95\textwidth]{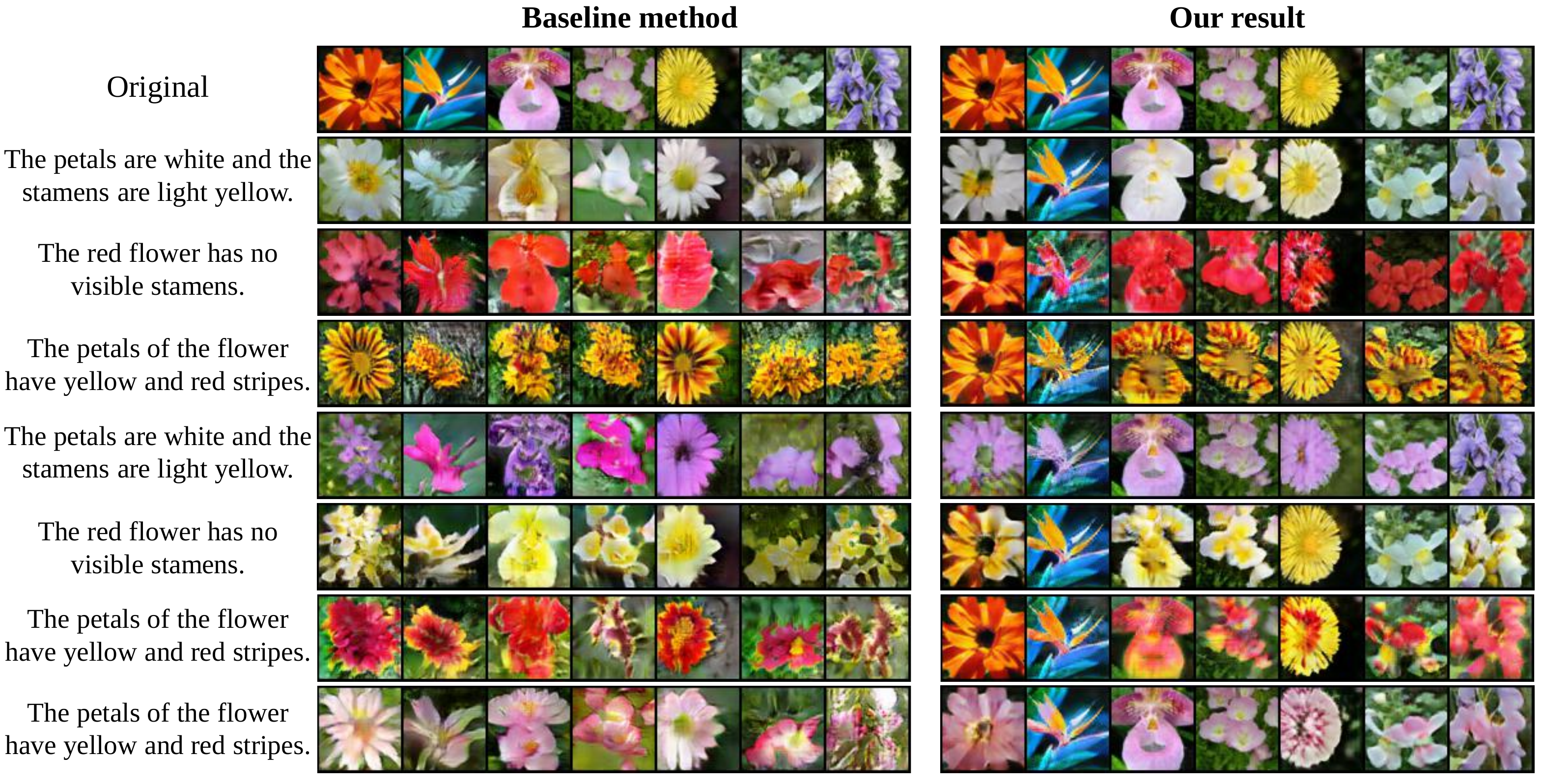}}}   
     \caption{Example results of the image synthesis problem in \cite{dong2017semantic}. The baseline method \cite{dong2017semantic} and our method (MC-GAN without mask) on Oxford-102 flower dataset} 
  \label{fig:trans_flower}%
  \end{figure}
  
\subsection{Interpolation and Variety}
\label{etc_exp}
\begin{figure}[b]
  \centering  
 {\bmvaHangBox{\includegraphics[width=0.95\textwidth]{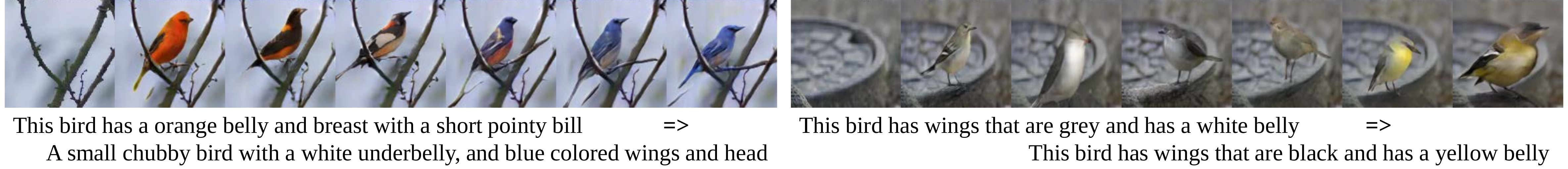}}}       
    \caption{Generated images using the interpolation of two text embeddings corresponding to the text descriptions below. The leftmost is the base image.}
  \label{fig:interpolation}%
  \vspace{-3mm}
\end{figure}

To generate images appropriately for various sentences, the latent manifold of the text embedding should be continuously trained. We generated continuously changing images by linearly interpolating the two different text embeddings from the sentences shown at the bottom of Fig. \ref{fig:interpolation}. The figure shows some example images that changes its color smoothly (mainly from orange to blue / from gray to yellow) following interpolated text embedding under the same noise and image conditions.

\begin{figure}[t]
  \centering  
    {\bmvaHangBox{\includegraphics[width=0.95\textwidth]{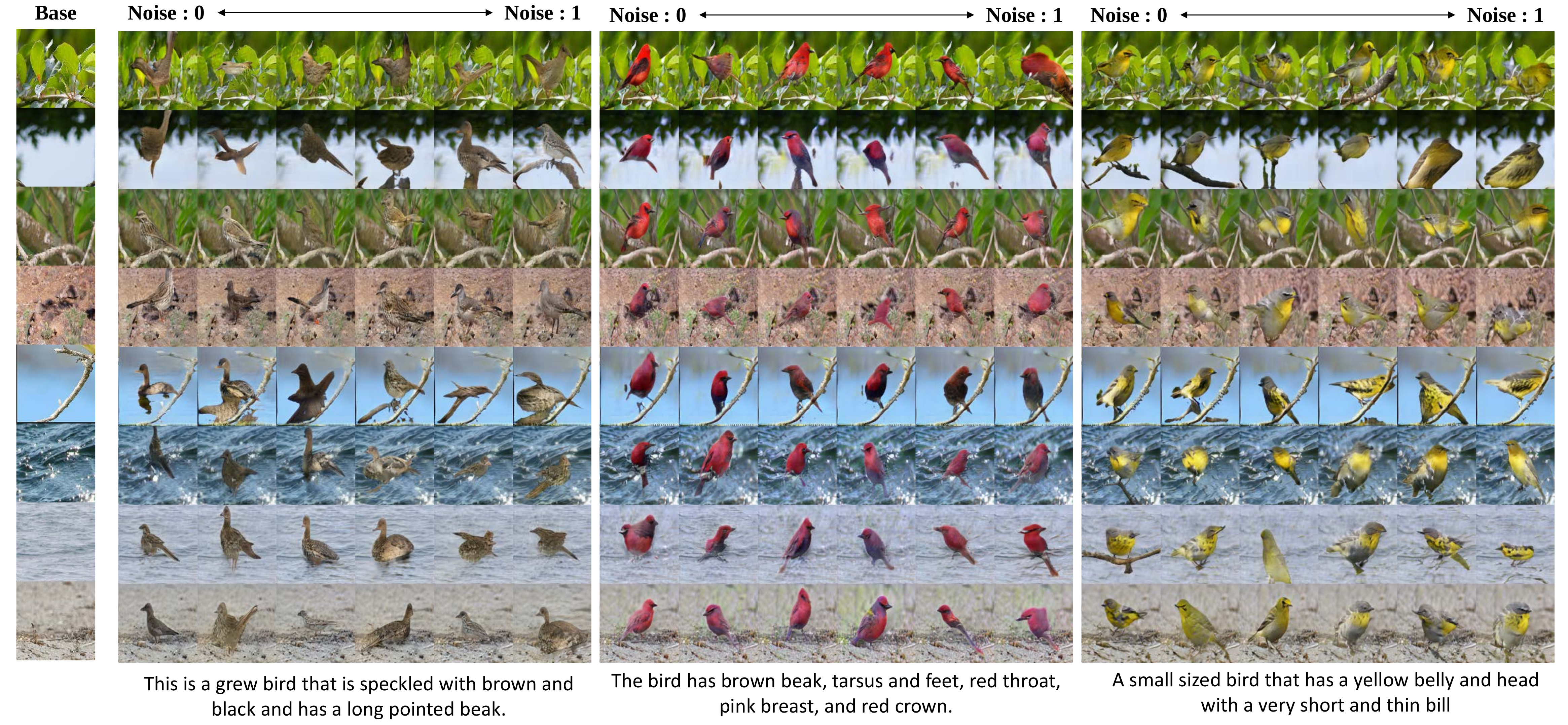}}}   
     \caption{Variety and stability of MC-GAN. The images are generated by linearly interpolating the noise vector $z$ from all-zero to all-one vector.}
  \label{fig:v_s}%
  \vspace{-2mm}
\end{figure}

As another experiment, we generated images using linearly interpolating the two noise vectors, which are all-zero ($z_0$) and all-one ($z_1$) vectors, under the same text and image conditions to demonstrate our model’s variety and stability. Fig \ref{fig:v_s} shows some resultant images.
Although it depends on the text and image conditions, but we usually got half of visually successful samples as can be seen in Fig \ref{fig:v_s}. 

\subsection{The effect of switch}
\label{swtich}
\begin{figure}
  \centering  
 {\bmvaHangBox{\includegraphics[width=0.95\textwidth]{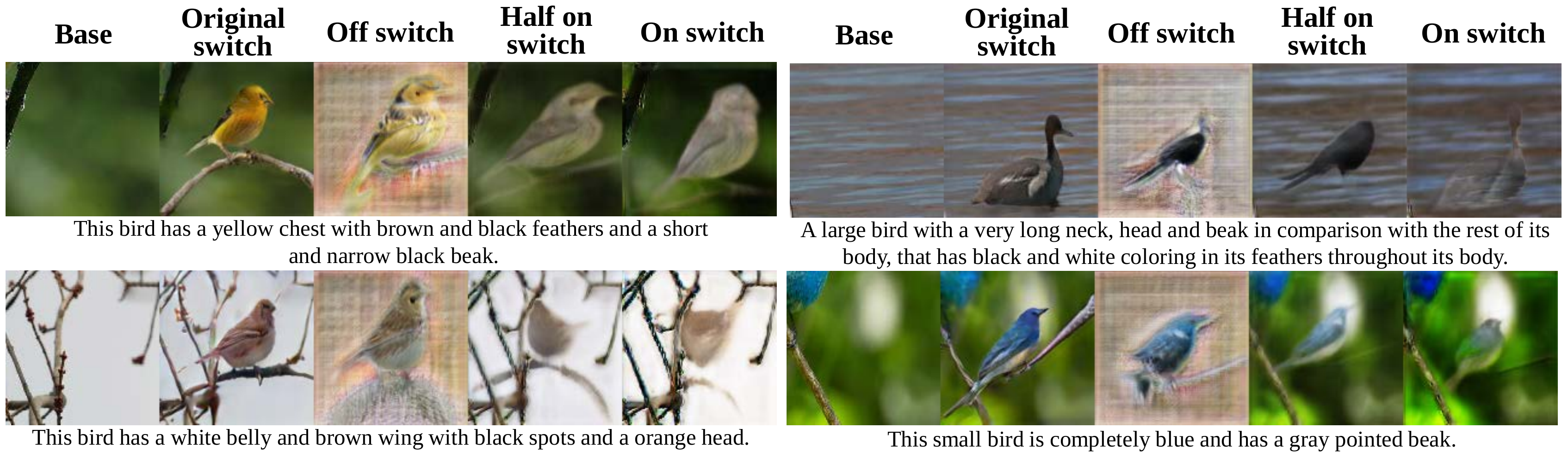}}}       
 	\vspace{-2mm}
    \caption{Experiment of generated images of switch transition in the Synthesis block}
  \label{fig:switch}%
  \vspace{-1mm}
\end{figure}

The switch in the synthesis block prevents the background and the foreground from overlapping each other. 
We compared the images generated by changing the switch value (the output value of the sigmoid function) under the fixed trained model of MC-GAN with the same base image, text and noise conditions. 
Fig. \ref{fig:switch} shows some results.
If we turn off all the switches (by zeroing the values) to prevent background information from being added, the original background disappears and only the generated object and the changed background are present. 
If all the switch values are set to 0.5 (half on), the background is reconstructed, but the object and the background slightly overlap and the image gets blurred compared to that of the original MC-GAN.
Finally, when all the switches are turned on, the original background information are added without suppression. In this case, the object and the background overlap with each other and the object is not properly visualized. 
By this experiment, we can verify that the switch in the synthesis block analyzes the current image and text conditions and adjusts the image feature flexibly to assist the proper synthesis of an image.

\subsection{ High Resolution Image Generation by Stacked GAN }
\label{resolution}
Based on the proposed MC-GAN, we additionally introduce a model to generate high resolution images by adding the StackGAN style two stage generator.
The initial spatial size of feature map in MC-StackGAN is $4\times4$ and two multiple GANs were stacked for generating $128\times128$ images like \cite{zhang2017stackgan,Han17stackgan2}.
The structure of the first GAN is the same as our original method and the second GAN takes a $64\times64\times64$ final feature map from the first GAN and a text embedding code from \textit{conditioning augmentation} without noise vector as mentioned in \cite{zhang2017stackgan,Han17stackgan2}. The first GAN's final feature map is concatenated with the replicated text embedding code.
We applied a convolution to the associated feature map with batch normalization and ReLU.
Finally, we used one more synthesis block and an upsampling layer to generate $128\times128$ images.
MC-StackGAN generates objects more stably than MC-GAN, but tends to transform the original base image compared to MC-GAN.

\begin{figure}[h]
  \centering  
    {\bmvaHangBox{\includegraphics[width=0.95\textwidth]{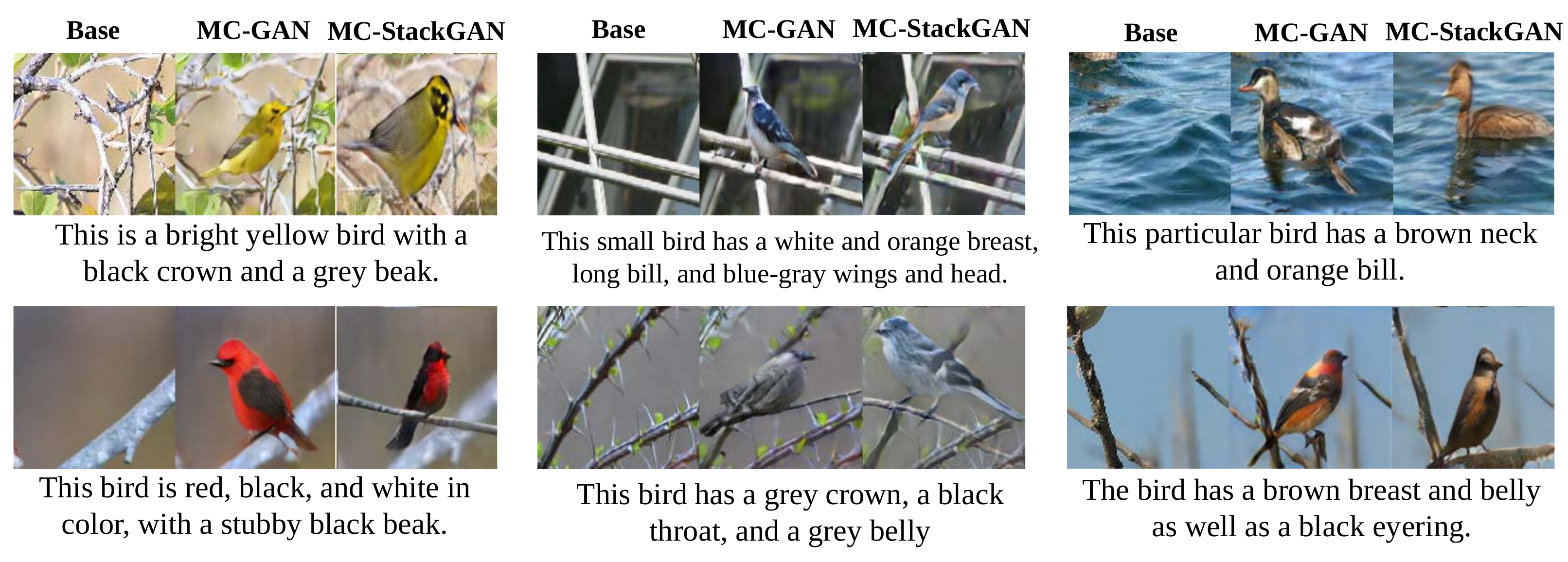}}}   
     \caption{Examples of MC-GAN and MC-StackGAN. From left to right: Base image, the synthesized image from MC-GAN $(128\times128)$, the synthesized image from MC-StackGAN $(128\times128)$ on Caltech-200 bird test dataset} 
  \label{fig:compare_bird}%
\end{figure}

\begin{figure}
  \centering  
    {\bmvaHangBox{\includegraphics[width=0.95\textwidth]{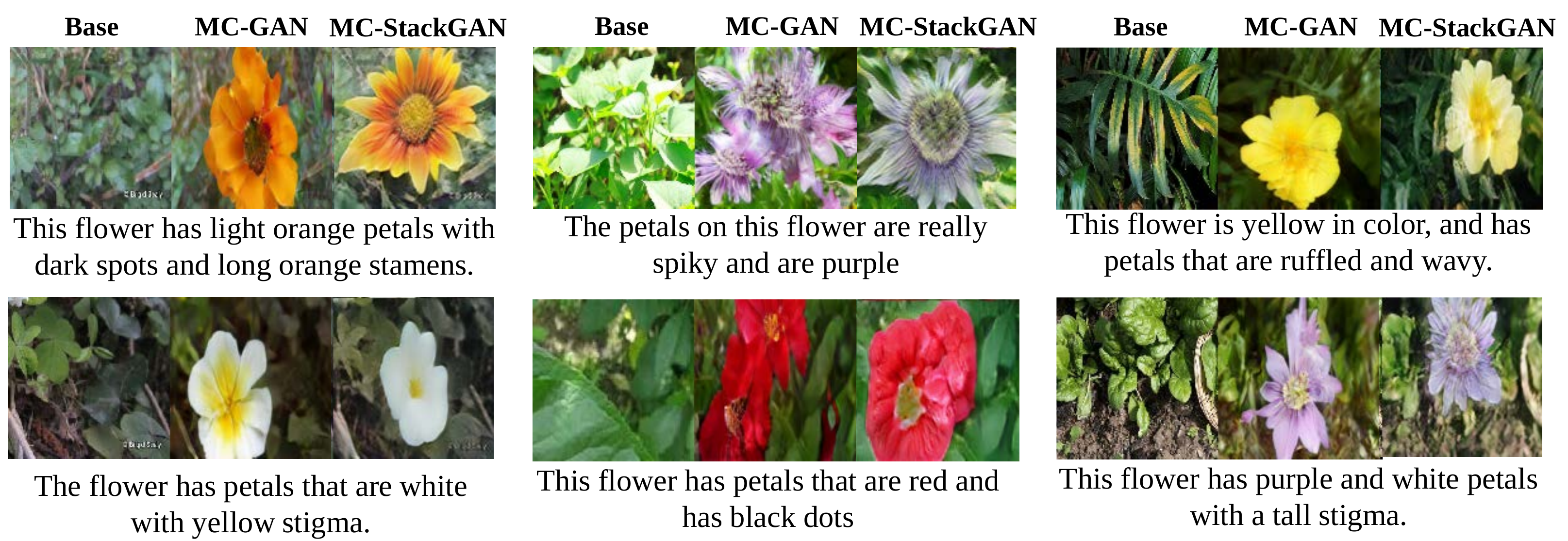}}}   
     \caption{Examples of MC-GAN and MC-StackGAN. From left to right: Base image, the synthesized image from MC-GAN $(128\times128)$, the synthesized image from MC-StackGAN $(128\times128)$ on Oxford-102 flower test dataset} 
  \label{fig:compare_flower}%
\end{figure}%

\vspace{-1mm}
\section{Conclusion}

In this paper, we introduced a new method of GAN to generate an image given a text attribute and a base image.
Different from the existing text-to-image synthesis algorithms only considering the foreground object, the proposed method aims to generate the proper target image, as well as preserving the semantics of the given background image.
To solve the problem, we newly proposed a MC-GAN structure and a synthesis block which is a core component enabling a photo-realistic synthesis by smoothly mixing foreground and background information.  
Using the proposed method, we confirmed that our model can generate diverse forms of a target object according to the text attribute while preserving the information of the given background image.
We also confirmed that our model can generate the object even when the background image does not include the same kind of object as the target, which is difficult for existing works.

\section{Acknowledgement}
\label{ack}
 This work was supported by Green Car Development project through the Korean MTIE (10063267) and Next-Generation Information Computing Development Program through the National Research Foundation of Korea (2017M3C4A7077582)


\bibliography{MCGAN}

\end{document}